# Graph-Theoretic Measures for Interpretable Multicriteria Decision-Making in Emergency Department Layout Optimization

**Ola Sarhan[a], Manal Abdel Wahed[a], Muhammad Ali Rushdi[a,b*]**

[a] Biomedical Engineering and Systems, Cairo University, Giza 12613, Egypt
[b] School of Information Technology, New Giza University, Giza 12256, Egypt

## Abstract

Overcrowding in emergency departments (ED) is a persistent problem exacerbated by population growth, emergence of pandemics, and increased morbidity and mortality rates. Thus, automated approaches for ED layout design have recently emerged as promising tools for boosting healthcare service quality. Still, ED design typically involves multiple conflicting objectives, where the interpretability of the associated solutions depends on the availability of intuitive metrics that can capture ED layout complexity. In this paper, we propose graph-theoretic measures to evaluate and rank ED layouts produced by a multi-objective metaheuristic optimization framework with the non-dominated sorting genetic algorithm (NSGA-II) and generalized differential evolution (GDE3). Indeed, Pareto-optimal ED layouts were sought to minimize patient flow cost while maximizing closeness between ED service areas. Then, the layouts were evaluated based on local graph measures (degree centrality, betweenness, clustering coefficient, closeness centrality, nodal strength, and eccentricity) as well as global ones (global efficiency, network characteristic path length and transitivity). Then, a multi-criteria decision-making technique was employed to rank the layouts based on either the objective functions, the graph measures, or combinations of both. The ranking results on a real-world scenario show that the top-ranking layouts are the ones with the best graph-theoretic values. This shows that the graph-theoretic measures can enhance solution interpretability and hence help medical planners in selecting the best layouts. In comparison with the input layout, optimal NSGA-II and GDE3 solutions reduce the patient flow cost by 18.32% and 11.42%, respectively. Also, the two solutions improve the closeness by 14.5% and 18.02%, respectively.

---

* Corresponding author:

Email addresses: mrushdi@eng1.cu.edu.eg

# 1. Introduction

The emergency department (ED) is a crucial component of any healthcare facility, providing immediate medical care to patients with urgent healthcare needs. It is often the first point of contact for individuals experiencing medical emergencies, trauma, or sudden onset of severe symptoms. However, emergency care systems have been strained by several problems including most recently the coronavirus [1], cholera [2], and armed conflicts [3]. Such problems have put pressure on emergency care systems especially under low-resource settings of low- and middle-income countries (LMICs) [4]. Emergency department (ED) visits surged by 41% between 1995 and 2009, jumping from 96.5 million to 136.1 million. This significant rise in demand highlights the growing complexity of the ED systems. Surprisingly, however, the number of hospitals EDs itself decreased by 27% during this same period, going from 2,446 to 1,779 [4]. The high volume of patients coupled with the need for immediate treatment mandates careful attention to emergency department design considerations.

Effective ED design and planning is crucial for improving healthcare delivery and service quality. Recent breakthroughs in computer technology have empowered designers and planners with powerful new tools to create and evaluate a wider range of ED layout options with greater accuracy and understanding [5]. The optimization of healthcare facility layouts is a complex and challenging task, that strives to improve workflow efficiency and patient satisfaction. Different methodologies and approaches have been used in the literature to solve emergency department layout problems (EDLPs) via mathematical modelling, simulation, and optimization [6]. The EDLPs typically involve multiple conflicting objectives related to patient flow, closeness among service areas, material handling costs, etc. Moreover, several quality metrics have been proposed for evaluating the ED layouts. Still, there is a still need for alternative metrics that can intuitively capture the increasing complexity of modern healthcare facilities.

In this paper, we propose novel graph-theoretic measures to evaluate ED facility layouts produced by a multi-objective metaheuristic optimization framework. In particular, sets of Pareto optimal ED layouts were sought to decrease patient flow and increase closeness between the ED service areas. Then, a multi-criteria decision-making technique was employed to rank the candidate layouts according to different criteria on the objective functions, the graph measures, or combinations of both.

In Section 2, we give a brief introduction on the hospital facility layout problems (Section 2.1) and then explore the literature about the approaches for layout generation (Section 2.2). In Section 2.3, graph-theoretic notions and measures are reviewed, while multi-criteria decision-making approaches for layout evaluation and selection are reviewed in Section 2.4. In Section 3, the formulation of our ED design problem, the associated mathematical model, multi-objective optimization techniques, and different evaluation techniques are explained. Finally, the experimental settings are given in Section 4 while the experimental results are illustrated in Section 5. Section 6 provides conclusions and recommendations for future work.

# 2. Literature Review

This section provides a concise review of literature on the hospital facility layout design problem. The review is divided into four parts: a brief introduction about hospital facility layout problem, conventional approaches for layout generation, graph-theoretic measures of layout plans, and multi-criteria decision-making approaches for layout evaluation.

### 2.1. Hospital facility layout problem

Planning or reconfiguring a layout of a hospital may be a complex task and therefore the development of quantitative planning techniques has been considered since the late 1970s [7]. Confusing layouts can increase patients' anxiety [8] and the associated patient flow instability negatively impact healthcare workflows [9], [10]. Also, length-of-stay variations and patient transfer from inpatient to outpatient care may cause uncertainties in patient flow [11]. To address these challenges, a hospital facility layout problem (FLP) involves the arrangement of hospital units in certain locations to satisfy design requirements and optimize some objectives [12]. Any FLP can be either a static FLP (where the patient and material flows between units remain constant throughout the planning horizon [6]), or a dynamic FLP (where the flows vary over the planning horizon [13]). While most of the FLPs were of the static type [14], [15], few studies considered dynamic FLPs [16]. Moreover, FLPs can be divided into single-row FLP, double-row FLP, multi-row FLP, loop FLP, and open-field LP [6]. A FLP can involve [6] minor changes, unit rearrangement, unit relocation in an existing layout, or creating a new layout to optimize certain objectives (e.g., minimizing patient flow costs, travel distances between departments, or material handling costs) [17].

### 2.2. Approaches for layout generation

Any type of FLP is solved by either using mathematical models, experts' knowledge, or computer-aided planning tools. Mathematical models are the most popular tools in the literature that can achieve optimal sets of solutions [6]. Most commonly, a hospital FLP can be simplified and formulated as a quadratic assignment problem (QAP) [18]. El Shafei [7] formulated the hospital design layout problem (HDLP) as a QAP and proposed a heuristic solution, where the objective is to minimize the patient travel distance. Butler *et al.* [19] introduced a two-phase HDLP approach that seeks to allocate beds to services and determine the relative service locations. The first phase used a quadratic integer programming model for bed and service allocation. In the second phase, the generated layout was evaluated via a simulation model. Yeh [20] investigated the allocation of services in a hospital building using an annealed neural network. The allocation problem was formulated as a QAP, for which the objective was a linear combination of the layout distances, neighboring preferences, and constraint violations. Cubukcuoglu *et al.* [21] introduced a new hospital layout design approach that leverages QAP to optimize space planning within existing facilities. This method focused on minimizing the flow of people and materials within the hospital.

Other layout design approaches involve formulating a FLP as a mixed-integer programming problem with a distance-based objective [17]. Chraibi *et al.* [22] formulated the operating theater layout problem as a mixed-integer linear program, which was subsequently solved using the CPLEX mathematical programming solver. Arnolds and Nickel [23] investigated hospital ward layout planning, proposed linear program formulations, and obtained solutions using CPLEX. Gai *et al.* [24] proposed an integrated method to select a proper healthcare department layout with the minimum operating cost and maximum system efficiency.

Furthermore, approximate and heuristic methods have been proposed for solving FLPs. Such methods can be categorized into construction, improvement, or metaheuristic methods [13]. Metaheuristic methods [6] have been most commonly and effectively used, and they include genetic algorithms, differential evolution and tabu search. For instance, Liang *et al.* [25] presented a multi-search tabu-search (TS) algorithm for FLPs. Emami *et al.* [26] introduced a multi-objective optimization formulation for a dynamic layout problem, which was solved using the NSGA-II algorithm, differential evolution, and Pareto-simulated annealing. Zuo *et al.* [14] optimized an emergency department layout using multi-objective tabu search. Huo *et al.* [15] presented a novel multi-objective double-row model suitable for a multi-floor HFLP of a working hospital using the NSGA-II algorithm with adaptive local search.

Furthermore, graph theory has been occasionally used in architectural planning and design. For example, Foulds [27] employed graph-theoretic heuristics for the plant layout problem. Roth and Hashimshony [28] developed graph-theoretic models for architectural design and decomposed complex graphs using the max-flow min-cut algorithm. Huang *et al.* [29] introduced an automatic graph-theoretic method for representing spatial information in CAD plans, where geometric primitives were analyzed, key points were identified as graph nodes, while walls and openings corresponded to graph edges. Assem *et al.* [30] proposed a graph-theoretic approach for scoring and evaluating operating theatre layouts. For a given layout, an adjacency matrix representation is constructed to show the closeness among different areas. The spiral technique is then used to build a layout block plan from the adjacency matrix by prioritizing spaces based on their importance to other areas, where higher importance spaces are placed more centrally. Finally, the resulting layout is scored based on how well it meets adjacency needs. This scoring method helps optimize OT design for better workflow. Arnolds *et al.* [31] provided an extensive review on layout problems in healthcare and introduced a graph-theoretic approach for designing a large hospital layout. Donato *et al.* [32] employed graph-theoretic concepts into building information modeling (BIM) for preliminary analysis of layout distribution, and they applied their approach to design the layout of a surgical department in an Italian hospital. Lakshmi *et al.* [33] investigated different graph-theoretic approaches for architectural design and urban planning. Reychav *et al.* [34] offered a novel graph-theoretic technique to yield insights from the analysis of a hospital's ED data collection. The collected data is used to create a time varying graph (TVG), a type of graph where connections and properties can change over time. In this case, the nodes represent patients and ED service areas, while edges represent transitions between them (e.g., moving from triage to consultation). The timestamps reflect the timing of each transition (a time-tree graph). Several graph metrics are employed to analyze patient flow within the ED such as degree centrality (to retrieve the busiest unit), shortest paths (to find the length of stay for patient in ED) and temporal congestion points. Bisht *et al.* [35] introduced G2PLAN, a mathematical software tool that transforms graph representations into floor plans that satisfy user-specified constraints. This tool can generate thousands of layouts within milliseconds, while maintaining high reliability and effectiveness. Amiri Chimeh and Javadi [36] introduced a multi-objective formulation for generating healthcare department layouts using planar adjacency graphs. Hassanain *et al.* [37] presented a two-pronged approach for evaluating healthcare facilities based on assessing adherence to standards and analyzing layout efficiency through graph theory. In the graph models, the nodes represent departments, while the edges represent connections (adjacency relationships) between them. A layout score was determined using a graph heuristic algorithm. This score reflects the efficiency of patient flow and departmental connectivity within the facility.

### 2.3. Graph-theoretic measures for layout evaluation

Graph-theoretic evaluation metrics have been introduced in several applications in computational neuroscience [38], computer-aided diagnosis [39], and traffic flow prediction [40]. Rubinov *et al.* [38] discussed the construction of brain networks from neurological data and described different measures of structural and functional network connectivity to reveal neurological disorders. Hu *et al.* [40] predicted the flow of traffic by using complex network measures based on spatial-temporal information between stations to analyze the importance and the degree of mutual influence of nodes in a network.

### 2.4. Approaches for layout evaluation and selection

Assessment and evaluation of potential hospital layouts are crucial for identifying the best layout. Despite the progress in generating layout alternatives, the field of facility layout design currently lacks advanced methods for effectively evaluating and ranking such alternatives [6]. Nevertheless, common methodologies for layout evaluation include multicriteria decision-making (MCDM) methods [6], fuzzy constraint theory [41], data envelopment analysis [42], or non-linear programming models [6]. Some of the most commonly used MCDM methods are elimination et choix traduisant la realité (ELECTRE) [43], a technique for order of preference by similarity to ideal solution (TOPSIS), analytic network process (ANP), analytic hierarchy process (AHP), preference selection index (PSI), and simple additive weighting (SAW).

Aiello *et al.* [44] employed multi-objective constrained genetic algorithms to obtain a set of Pareto solutions for hospital layout design. Then, the ELECTRE method was used to detect the best solution across the whole set of solutions. Shafii *et al.* [45] aimed to evaluate the service quality of teaching hospitals in Iran using the TOPSIS and fuzzy AHP methods. Hassanain *et al.* [46] investigated the process of redesigning old healthcare facilities to meet international standards, by applying a fuzzy TOPSIS technique for layout evaluation and ranking. A reallocation algorithm was employed based on graph heuristics techniques and a bubble plan.

In our work, we further explore the potential of graph-theoretic metrics for evaluating medical planning outcomes. We claim that an analogy can be established between brain connectivity networks [38] and hospital layout organization. As the brain connectivity networks consist of brain regions connected by anatomical tracts or by functional association, hospitals comprise networks of departments connected by relationships dictated by patient flow, standards, expert knowledge, and stakeholders' opinions. On the basis of this analogy, we propose novel graph-theoretic measures to evaluate ED facility layouts produced by a multi-objective metaheuristic optimization framework. Specifically, Pareto optimal ED layouts are generated such that patient flow is minimized while closeness between ED service areas is maximized. Then, a MCDM technique is exploited to rank the top candidate ED layouts according to different criteria based on the objective functions, the introduced graph measures, and their combinations.

In summary, the main contributions of this paper are of three folds:

1. Reviewing the clinical design standards of emergency departments (ED) and formulating the ED layout design problem as a multi-objective mixed optimization problem. For solving this problem, meta-heuristic algorithms were employed and compared to generate Pareto optimal solutions.

2. Introducing graph-theoretic measures for evaluating the relative strengths and weaknesses of the Pareto optimal ED layouts from the point of view of complex networks.
3. Employing multi-criteria decision-making techniques to fuse conventional measures, graph-theoretic metrics, and expert opinions in order to effectively rank the Pareto ED layouts.

# 3. Material & Methods

## 3.1. *Emergency Department Layout Problem (EDLP)*

A real-world EDLP was formulated by Zuo et al.[14] for an existing hospital in Dalian, China, as shown in Fig. 1. The layout consists of 20 service areas arranged in three rows and separated by two horizontal corridors. Instead of the nomenclature suggested by Zuo *et al.* [14], we used revised names for the service areas following international and reference standards [47] including the space planning guides of the US Department of Defence [48], the US Department of Veteran's Affairs [49], the Irish Standards for Emergency Department Design and Specifications [50], and the Egyptian Design Standards for Hospitals and Medical Facilities [51]. Table 1 lists the 20 service areas with their revised names, the dimensions of each area, and whether each area is fixed or relocatable. Service areas 13-19 (highlighted in dark grey) were fixed based on expert opinions, while the service areas 0-12 (highlighted in light grey) are relocatable and their placement shall be decided through a numerical optimization approach. The shaded area in the upper row (namely, the "Airborne Infection Isolation area") is relatively fixed and it may be moved slightly to the left or right. To do this, there should be 4 service areas on the left of this area. The "toilet" is fixed, while the reserved area is to be used in the future. In the lower row, the "security room" can be slightly moved to the left or right, while the "doctors' office" is fixed. The width of the vertical corridors can be changed within limits, but it mustn't be less than 2.4 m according to international standards [51]. The service areas in each row must be located within certain ranges (indicated by $L_1$, $L_2$, and $L_3$ in Fig. 1). The dimensions of each corridor and the length of each row are provided in Table 2 along with other EDLP parameters.

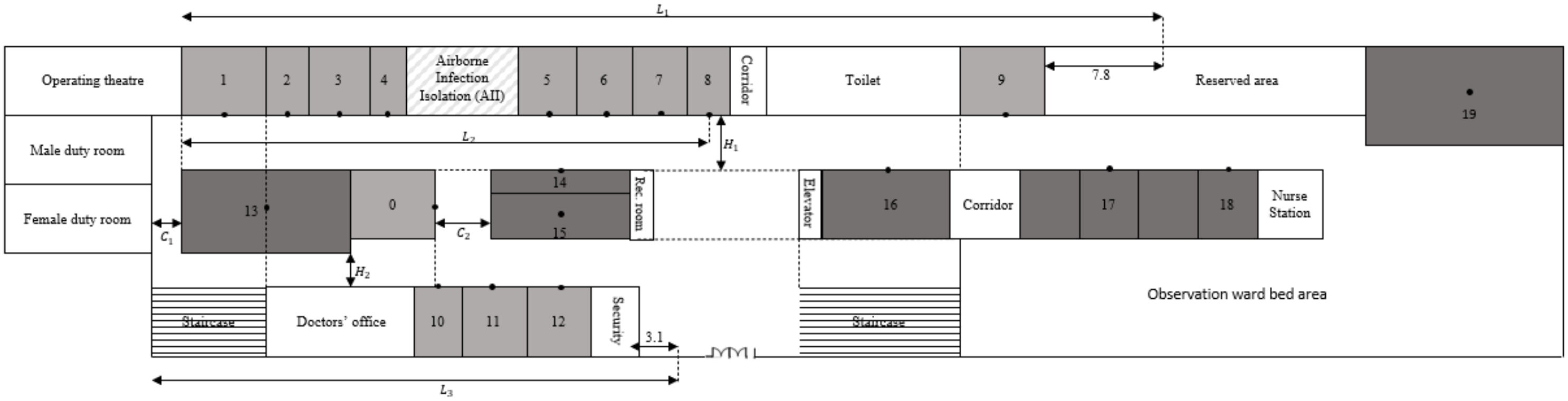

*Figure 1 A real-world ED layout in a hospital in Dalian, China; $C_1$=2.4m & $C_2$=3.6m [5].*

## 3.2. *Multi-objective EDLP*

### *3.2.1. Objective functions*

The EDLP typically has two optimization objectives: patients' flow cost and closeness of service areas. Using the 2D spatial locations of the service areas, these two objectives can be calculated as follows.

1) <u>Patients' Flow Cost ($F_1$):</u> The cost of patient flow between any two service areas is directly proportional to the travel distance and the travel frequency (the number of patients' trips between the two areas). Thus, the minimization of this cost can be mathematically formulated as:

$$\min_{\substack{x_i, y_i, x_j, y_j \\ 0 \le i,j \le N-1}} F_1 = \sum \sum f_{ij} d_{ij}, \tag{1}$$

$$d_{ij} = |x_i - x_j| + |y_i - y_j|, \tag{2}$$

where $f_{ij}$ represents the number of patients flowing from the $i^{th}$ to the $j^{th}$ service areas during a certain fixed time, $d_{ij}$ is the rectilinear distance between the two service areas, and $x_i$ and $y_i$ are respectively the $x$-coordinate and y-coordinate of the entrance (or centroid) of the $i^{th}$ service area.

Note that the design variables are the coordinates of the relocatable service areas only. To minimize $F_1$ any two service areas with high travel frequency are desired to be closer to each other. The pairwise travel frequencies (i.e., the patient flow costs) for the 20 service areas in the investigated ED example are presented in Table 3.

*Table 1 Service area information*

| No. | Service Area/Unit | $Wi$ (m) | $Di$ (m) | Fixed |
|---|---|---|---|---|
| 0 | Soiled utility room | 5.5 | 4.5 | No |
| 1 | Gynecology & Obstetrics emergency | 5.5 | 4.5 | No |
| 2 | Ophthalmic emergency | 2.75 | 4.5 | No |
| 3 | ENT emergency | 3.95 | 4.5 | No |
| 4 | Dental emergency | 2.4 | 4.5 | No |
| 5 | Neurological emergency | 3.6 | 4.5 | No |
| 6 | Dermatology emergency | 3.6 | 4.5 | No |
| 7 | Surgical emergency | 3.6 | 4.5 | No |
| 8 | Orthopedic emergency | 2.75 | 4.5 | No |
| 9 | Clinical laboratory | 5.5 | 4.5 | No |
| 10 | Portable imaging room | 2.75 | 4.5 | No |
| 11 | Blood bank | 4.5 | 4.5 | No |
| 12 | Pediatric emergency | 4.5 | 4.5 | No |
| 13 | Operating room | 11 | 5.4 | Yes |
| 14 | Triage (1) | 9.1 | 1.5 | Yes |
| 15 | Registration | 9.1 | 3 | Yes |
| 16 | Medication room | 6 | 4.5 | Yes |
| 17 | Triage (2) | 3.9 | 4.5 | Yes |
| 18 | Emergency transfusion | 3.9 | 4.5 | Yes |
| 19 | Radiology | 13.6 | 6.4 | Yes |

1) <u>Service areas' closeness ($F_2$):</u>

For each pair of service areas $i$ and $j$, a closeness rating $r_{ij}$ was specified by medical planning experts. AEIOUX rating system [52] is used to determine these closeness ratings. It was defined as, A ≡ absolutely necessary which takes a value of 5, E ≡ especially important which takes a value of 4, I ≡ important which takes a value of 3, O ≡ ordinary closeness which takes a value of 2, U ≡ unimportant which takes a value of 1, X ≡ undesirable which takes 0 value.

*Table 2 EDLP main parameters*

| Parameter | Description |
|---|---|
| $N$ | Number of service areas |
| $D_i$ | Room depth of the $i^{th}$ service area ($0 \leq i \leq N-1$). |
| $W_i$ | Room width of the $i^{th}$ service area ($0 \leq i \leq N-1$). |
| $f_{ij}$ | Number of patients' trips from the $i^{th}$ to the $j^{th}$ service areas ($0 \leq i, j \leq N-1$). |
| $H_1$ | Depth of the upper horizontal corridor, $H_1$ = 3.6 m |
| $H_2$ | Depth of the lower horizontal corridor, $H_2$ = 2.4 m |
| $L_1$ | Maximum length of the upper row, $L_1$= 64 m |
| $L_2$ | Maximum length of the upper row, $L_2$ = 34.3 m |
| $L_3$ | Maximum length of the upper row, $L_3$ = 34.3 m |
| $C_1$ | Minimum width of $C_1$ = 2.4 m |
| $C_2$ | Minimum width of $C_2$= 3.6 m |

The values for all ratings are given in Table 4. Also, an adjacency coefficient $b_{ij}$ is used to measure the adjacency degrees between the service areas $i$ and $j$. These coefficients are calculated following the method in [14]. First, we calculate the average width over all relocatable service areas

$$W_a = \frac{1}{N} \sum_{0 \leq i \leq N-1} W_i \qquad (3)$$

where $N$ is the number of service areas. Second, we compute the normalized number of areas between the $i^{th}$ and $j^{th}$ service areas as

$$nd_{ij} = \frac{d_{ij}}{W_a} = \frac{|x_i - x_j| + |y_i - y_j|}{W_a} \qquad (4)$$

Finally, pairwise adjacency coefficients $b_{ij}$ are decided based on $nd_{ij}$ (rounded to the closest integer) as follows:

$$b_{ij} = \begin{cases} 1.0, & nd_{ij} = 0 \\ 0.8, & nd_{ij} = 1 \\ 0.6, & nd_{ij} = 2 \\ 0.4, & nd_{ij} = 3 \\ 0.2, & nd_{ij} = 4 \\ 0, & nd_{ij} \geq 5 \end{cases} \qquad (5)$$

Thus, the minimization of a closeness objective can be mathematically formulated as:

$$\min_{\substack{x_i, y_i, x_j, y_j \\ 0 \leq i,j \leq N-1}} F_2 = M - \sum\sum r_{ij}\, b_{ij} \qquad (6)$$

where $M$ is a constant number and set here to be 300 to be greater than output summation value of product of closeness rating and adjacency coefficient.

### *3.2.2 Constraints*

The following feasibility constraints on the ED dimensions (in meters) are enforced to adhere with clinical design guidelines:

$$L_1 \leq 64 \qquad (7)$$

$$L_2, L_3 \leq 34.3 \qquad (8)$$

$$2.4 \leq C_1 \leq 4 \qquad (9)$$

$$3.6 \leq C_2 \leq 4.6 \qquad (10)$$

*Table 3 Patients' flow cost ($f_{ij}$) for each pair of 20 service areas in the examined ED layout [14].*

| **Service Area/ Unit** | **0** | **1** | **2** | **3** | **4** | **5** | **6** | **7** | **8** | **9** | **10** | **11** | **12** | **13** | **14** | **15** | **16** | **17** | **18** | **19** |
|---|---|---|---|---|---|---|---|---|---|---|---|---|---|---|---|---|---|---|---|---|
| **0** | 0 | 287 | 223 | 688 | 961 | 5684 | 229 | 10908 | 5797 | 0 | 0 | 0 | 2133 | 0 | 27751 | 0 | 0 | 1878 | 0 | 0 |
| **1** | | 22 | 0 | 0 | 0 | 0 | 0 | 0 | 0 | 958 | 170 | 4 | 0 | 0 | 0 | 954 | 0 | 0 | 20 | 18 |
| **2** | | | 498 | 0 | 0 | 0 | 0 | 0 | 0 | 184 | 0 | 0 | 0 | 2 | 0 | 3177 | 0 | 0 | 44 | 632 |
| **3** | | | | 2694 | 0 | 0 | 0 | 0 | 0 | 1360 | 116 | 0 | 0 | 16 | 0 | 7882 | 20 | 0 | 1878 | 1298 |
| **4** | | | | | 309 | 0 | 0 | 0 | 1 | 262 | 12 | 0 | 0 | 10 | 0 | 2732 | 0 | 0 | 102 | 606 |
| **5** | | | | | | 9 | 0 | 0 | 0 | 0 | 6122 | 2 | 0 | 0 | 0 | 23782 | 2 | 4 | 3792 | 9162 |
| **6** | | | | | | | 0 | 0 | 0 | 0 | 82 | 2 | 0 | 0 | 0 | 1882 | 0 | 0 | 1550 | 12 |
| **7** | | | | | | | | 2 | 8 | 29186 | 2416 | 0 | 0 | 7 | 0 | 42851 | 26 | 4 | 0 | 15380 |
| **8** | | | | | | | | | 2299 | 2298 | 554 | 0 | 0 | 246 | 0 | 18938 | 42 | 0 | 544 | 17162 |
| **9** | | | | | | | | | | 300 | 0 | 0 | 1938 | 0 | 57678 | 150 | 0 | 1992 | 0 | 0 |
| **10** | | | | | | | | | | | 0 | 0 | 224 | 0 | 7613 | 0 | 0 | 332 | 0 | 0 |
| **11** | | | | | | | | | | | | 0 | 0 | 0 | 0 | 1 | 0 | 0 | 0 | 0 |
| **12** | | | | | | | | | | | | | 2 | 0 | 0 | 5934 | 0 | 0 | 0 | 726 |
| **13** | | | | | | | | | | | | | | 0 | 0 | 1 | 0 | 0 | 0 | 0 |
| **14** | | | | | | | | | | | | | | | | 129011 | 0 | 2 | 53150 | 16314 |
| **15** | | | | | | | | | | | | | | | | 0 | 0 | 4904 | 0 | 0 |
| **16** | | | | | | | | | | | | | | | | | 0 | 0 | 0 | 0 |
| **17** | | | | | | | | | | | | | | | | | | 52 | 326 | 222 |
| **18** | | | | | | | | | | | | | | | | | | | 0 | 0 |
| **19** | | | | | | | | | | | | | | | | | | | | 0 |

*Table 4 Pairwise closeness ratings ($r_{ij}$) for 20 service areas in the examined ED layout [14].*

| Service Area | 0 | 1 | 2 | 3 | 4 | 5 | 6 | 7 | 8 | 9 | 10 | 11 | 12 | 13 | 14 | 15 | 16 | 17 | 18 | 19 |
|---|---|---|---|---|---|---|---|---|---|---|---|---|---|---|---|---|---|---|---|---|
| 0 | | 4 | 4 | 4 | 4 | 4 | 4 | 4 | 4 | 4 | 4 | 4 | 4 | 4 | 4 | 4 | 3 | 3 | 3 | 3 |
| 1 | | | 2 | 2 | 2 | 2 | 5 | 3 | 0 | 4 | 4 | 4 | 5 | 3 | 4 | 4 | 3 | 3 | 3 | 4 |
| 2 | | | | 5 | 5 | 2 | 0 | 2 | 1 | 4 | 4 | 4 | 2 | 3 | 4 | 4 | 3 | 3 | 3 | 4 |
| 3 | | | | | 5 | 2 | 0 | 2 | 1 | 4 | 4 | 4 | 2 | 3 | 4 | 4 | 3 | 3 | 3 | 4 |
| 4 | | | | | | 2 | 0 | 2 | 1 | 4 | 4 | 4 | 2 | 3 | 4 | 4 | 3 | 3 | 3 | 4 |
| 5 | | | | | | | 2 | 5 | 1 | 4 | 4 | 4 | 2 | 3 | 5 | 4 | 3 | 3 | 3 | 4 |
| 6 | | | | | | | | 3 | 0 | 4 | 4 | 4 | 5 | 3 | 4 | 4 | 3 | 3 | 3 | 4 |
| 7 | | | | | | | | | 3 | 4 | 4 | 4 | 5 | 3 | 4 | 4 | 3 | 3 | 3 | 4 |
| 8 | | | | | | | | | | 4 | 4 | 4 | 2 | 4 | 4 | 4 | 3 | 3 | 3 | 4 |
| 9 | | | | | | | | | | | 5 | 5 | 4 | 4 | 4 | 4 | 3 | 3 | 3 | 4 |
| 10 | | | | | | | | | | | | 5 | 4 | 4 | 4 | 4 | 3 | 3 | 3 | 4 |
| 11 | | | | | | | | | | | | | 4 | 4 | 4 | 4 | 3 | 3 | 3 | 4 |
| 12 | | | | | | | | | | | | | | 4 | 4 | 4 | 3 | 3 | 3 | 4 |
| 13 | | | | | | | | | | | | | | | 4 | 4 | 3 | 3 | 3 | 4 |
| 14 | | | | | | | | | | | | | | | | 4 | 3 | 3 | 3 | 4 |
| 15 | | | | | | | | | | | | | | | | | 3 | 3 | 3 | 4 |
| 16 | | | | | | | | | | | | | | | | | | 3 | 3 | 4 |
| 17 | | | | | | | | | | | | | | | | | | | 3 | 4 |
| 18 | | | | | | | | | | | | | | | | | | | | 4 |
| 19 | | | | | | | | | | | | | | | | | | | | 0 |

### 3.3. *Multi-objective meta-heuristic optimization algorithms*

We adopted two main algorithms to solve and optimize our EDLP. The two algorithms are metaheuristics optimization techniques: genetic algorithms, and differential evolution. These algorithms were chosen for reasons of being simple, globally optimal, and effective in dealing with noisy environments. However, like other global optimization techniques, these two methods are computationally expensive.

#### *3.3.1 Non-dominated Sorting Genetic Algorithm II (NSGA-II)*

The non-dominated sorting genetic algorithm II (NSGA-II) can be used to optimize the two objectives in (Eq. (1) and (6)). Given a population of potential solutions, conventional genetic algorithms use crossover and mutation parameters to rank individuals, select the best-fit ones, and hence generate a new offspring [53]. For the NSGA-II variant, ranking of individuals solutions is based on tournaments. Non-dominated ranking is done first to select and rank the best fronts. If two individuals are in the same front, a crowding distance is applied to choose the higher one. A pseudocode description of the NSGA-II technique is given in Algorithm A.I.

#### *3.3.2 Generalized Differential Evolution, Third Version (GDE3)*

The third version of the GDE algorithm [54] utilizes differential evolution operators, specifically mutation and crossover, to generate new candidate solutions. Besides the selection, GDE3 employs an archive-based selection mechanism to maintain a set of non-dominated solutions. The archive stores the best solutions found so far, and new offspring solutions are compared against the archive to determine their dominance status. The dominance characteristics influence both the mutation and crossover operations, allowing the GDE3 algorithm to adapt its behavior based on the given problem and current population. Consequently, the archive facilitates the preservation of diverse and high-quality solutions [54]. The description of GDE3 technique is given in Algorithm A.II.

### 3.4. *Layout design evaluation with graph-theoretic measures*

Graph theory can be used to describe relations between different functional service areas in a hospital layout. In this section, we review different graph-theoretic measures that evaluate different aspects of functional integration, segregation, and centrality of hospital layout designs. A complex system can be modeled by a graph or network representation with a set of nodes (vertices) and links (edges) between pairs of nodes. In our framework, *nodes* represent emergency department service areas, while *links* represent spatial connections (among adjacent service areas), functional connections (e.g., patient flow between service areas), or circulations (e.g., going through lab work several times). Links can also be associated with weight or directionality information. Binary links denote the presence or absence of connections, while weighted links additionally indicate connection strengths. Weights may represent patient flow density, transportation, and handling costs, communication efficiency, or functional closeness. Incorporating such weights can result in different graph-theoretic measures. For example, the inter-nodal distance in an unweighted graph is based on counting the number of edges between two nodes, while the distance in a weighted graph is the sum of the weights of the edges on the path between two nodes [38]. Directionality can also be considered to account for constraints of one-way movements between ED service areas. Such constraints help avoid potential crowding and streamline patient flow.

In this paper, we analyze the network connectivity patterns and the degree of mutual influence among ED service areas. This analysis can be carried out based on local or global graph metrics as explained next.

Among these metrics, the global efficiency, network characteristic path length (NCPL), and node strength, used weighted adjacency matrix of the graph.

*3.4.1. Local graph metrics*

Local measures include degree (D), degree centrality (DC), betweenness centrality (BC), closeness centrality (CC), clustering coefficient (C), eccentricity (ECC), and nodal strength (S).

*1- Degree & degree centrality*

The degree of an individual node is equal to the number of links connected to that node, which in practice can indicate the number of service areas connected to a specific ED service area. Degree restrictions vary from one area to the other. For example, an area with desired high accessibility (such as the triage) must have a high degree, while the degree of an airborne infection isolation unit would be low. Consequently, the individual values of the node degrees reflect the importance of the corresponding nodes and this in turn affects network resilience. The degree of the $i^{th}$ node in a graph can be mathematically defined as

$$D_i = \sum_{i,j=1}^{N} a_{ij} \qquad (11)$$

where $a_{ij}$ represents a binary value reflecting the adjacency between two nodes $i$ and $j$. If there is a link between nodes $i$ and $j$, $a_{ij} = 1$. Otherwise, $a_{ij} = 0$. Degree centrality is defined as the node degree divided by the maximum possible degree (i.e., n-1):

$$DC_i = \frac{D_i}{N-1} \qquad (12)$$

*2- Closeness centrality*

Closeness centrality for the $i^{th}$ node of a graph is defined as the reciprocal of the sum of the shortest distances between the $i^{th}$ node and all other nodes:

$$CC_i = \frac{N-1}{\sum_{i,j=1}^{N} \alpha_{ij}} \qquad (13)$$

where $\alpha_{ij}$ is shortest distance between two nodes $i$ and $j$ in an unweighted graph. The larger the value of the closeness centrality, the closer the node to the network center.

*3- Normalized betweenness centrality*

Betweenness centrality represents the degree to which nodes stand between each other. For example, in an emergency department layout, an area with a higher betweenness centrality would have more influence over the network, because more connections will pass through that area. In particular, this type of centrality indicates how often we can pass by a certain node during our movement through the shortest path between two random nodes in the network. Let $BC_i$ denote the ratio between the number of shortest paths through the $i^{th}$ node to the number of all shortest paths in the network:

$$BC_i = \sum_{h,j \epsilon N, h \neq j, h \neq i, j \neq i} \frac{\rho_{hj\ (i)}}{\rho_{hj}} \qquad (14)$$

where $\rho_{hj}$ is the number of shortest paths between $h$ and $j$, and $\rho_{hj}\ (i)$ is the number of shortest paths between $h$ and $j$ that pass through $i$. The normalized betweenness centrality for the $i^{th}$ node is defined as the betweenness centrality divided by the maximum number of $h$ and $j$ combinations ($h$ can have $N-1$ values excluding that of the $i^{th}$ node while $j$ can have $N-2$ values excluding those of the $i^{th}$ and $h^{th}$ nodes):

$$NBC_i = \frac{BC_i}{(N-1)(N-2)} \qquad (15)$$

*4- Clustering coefficient*

The clustering coefficient of the $i^{th}$ node of a graph is the fraction of the actual number of triangles around this node to the maximum number of triangles around that node:

$$C_i = \frac{E_i}{\binom{D_i}{2}} = \frac{2 \times E_i}{D_i(D_i-1)} \qquad (16)$$

where $E_i$ represents the actual number of edges between the neighbors of node $i$. In other words, this coefficient is equal to the fraction of the node's neighbors that also are neighbors of each other.

This coefficient represents the degree of aggregation for each node in the network. Highly clustered service areas are segregated from the rest of the ED and can work more efficiently and independently.

*5- Eccentricity*

For a weighted graph, the eccentricity of the $i^{th}$ graph node is the maximum distance between that node and any other graph node:

$$ECC_i = \max_{j \in N} ß_{ij} \qquad (17)$$

where $ß_{ij}$ is shortest distance between two nodes $i$ and $j$ in the weighted graph.

*6- Node strength*

For a weighted graph, the strength of a node is the sum of the weights of the edges to the neighboring nodes:

$$S_i = \sum_{e_{ij} \in E} w_{ij} \qquad (18)$$

The node strength is more informative than the node degree.

### *3.4.2. Global Graph metrics*

Global measures include global efficiency (GE), transitivity (T), and network characteristic path length (NCPL).

*1- Global efficiency (GE)*

It is the average inverse shortest path length. GE is inversely related to the characteristic path length. This feature reflects the efficiency of moving between different service areas in general, counter to the closeness centrality which is defined for each individual service area. When this feature is high, it means that it is easier to reach these service areas. Otherwise, when the length is infinite, efficiency is zero.

$$GE = \frac{1}{N} \sum_{i \in N} \frac{\sum_{j \in N j \neq i} ß_{ij}^{-1}}{N-1} \qquad (20)$$

*2-Transitivity*

It is the ratio of triangles to triplets, which represents the likelihood of a connection between service areas if they have a known mutual unit connected to both of them.

$$T = \frac{\sum_{i=1}^{N} 2t_i}{\sum_{i=1}^{N} D_i(D_i - 1)} \qquad (21)$$

where $t_i$ is the number of triangles around the node $i$.

*3-Network characteristic path length (NCPL)*

$$NCPL = \frac{1}{N} \sum_{i \epsilon N} \frac{\sum_{j \epsilon N, j \neq i} ß_{ij}}{N-1} \qquad (22)$$

### *3.4.3. Calculating the adjacency matrix*

Based on the graph measures introduced in Sections 3.4.1 and 3.4.2, several strategies have been explored for adjacency matrix calculation starting from the following general expression which combines the patient flow values and closeness ratings with a weighting parameter $\gamma$:

$$A_{ij} = \left( \gamma \, f_{ij} + (1 - \gamma) \, r_{ij} \right) d_{ij} \qquad (23)$$

For the explored strategies, $\gamma$ is varied follows:

Strategy (I) $(\gamma = 0)$: The service areas are grouped based on how closely they need to work together, while patient flow between service areas is discarded.

Strategy (II) $(\gamma = 0.25)$: Service area closeness relationships are prioritized over patient flow when organizing the service areas.

Strategy (III) $(\gamma = 0.5)$: Equal importance is placed on patient flow and service area closeness relationships.

Strategy (IV) $(\gamma = 0.75)$: The service areas are arranged primarily based on patient flow rather than the closeness relationships.

Strategy (V) $(\gamma = 1)$: Service areas are organized based on patient flow, positioning those with higher patient flow together, while the service area closeness relationships are discarded.

To choose the best strategy to follow, the adjacency matrices and the corresponding global graph measures for the Pareto solutions (obtained by both NSGA-II and GDE3) were calculated and normalized. The average value for each normalized graph measure is computed, and then the mean and standard deviation for each strategy is obtained. After that, the coefficient of variation $(Cv)$ is calculated as the ratio of the standard deviation to the mean:

$$Cv(l) = \frac{Std\ (l)}{mean(l)} \qquad (24)$$

where $l$ denotes the strategy number $(l = 1{:}\,5)$.

The strategy with the lowest value for the coefficient of variation is selected, as such a strategy exhibits the highest mean and the lowest standard deviation [36], [55].

## **3.5. *Optimal layout solution selection***

Multi-criterion decision making techniques can be employed to select the best layout solution[6] . In our work, the ELECTRE-III technique [56] is used. We preferred to use ELECTRE-III because this technique considers uncertainty of data into decision-making process [43], [44]. It can deal with qualitative data. It is found that the top-ranked alternatives by ELECTRE is same as ones obtained by other MCDM such as TOPSIS or SAW [43]. ELECTRE-III method is based upon a pseudo-criterion that accounts for the ambiguity and uncertainty in performance evaluation by applying appropriate thresholds. Thus, the ELECTRE-III technique can be used to select the best solutions based on the two conventional objective functions (Eq. (1), (6)), global graph measures, or the combination of conventional and graph-based criteria. A pseudocode description of the ELECTRE-III technique is given in Algorithm A.III. This technique has three types of thresholds. For each criterion $(i)$, ELECTRE-III selection thresholds have been calculated as percentages of the value of the respective objective functions. In particular, the indifference threshold $(q_i)$ was set to be less than or equal to 15%, while preference $(p_i)$ and veto $(v_i)$ thresholds were fixed as 30% and 50%, respectively. The weights of all objectives and/or global graph measures were all assumed to be equal 1 [44]. We used the ELECTRE-III implementation of the Python-based and Excel –XLSTAT package.

### 3.6. *Performance metrics and statistical significance tests*

Different metrics and statistical significance tests were used in our work to evaluate the performance of the layout optimization algorithms.

#### *3.6.1 Performance metrics*

We adopt the hypervolume, fitness value, timing, and set coverage (c-metric) to compare the solutions obtained by the NSGA-II and GDE3 methods[57]. The hypervolume (HV) is unary metric (it means that it can evaluate one of methods independently) [58] . It is a commonly used performance metric that measures the size of the objective space covered by an approximate Pareto-solution set. The higher the HV value, the better the quality of the Pareto solutions [59]. The average fitness value (FV) is the average of the objective fitness values computed over all Pareto solutions. The lower the FV, the better the solution quality. The set coverage (or c-metric) is a binary metric (it receives as parameter two approximation sets, $M_1$ and $M_2$, to be compared). C-metric of two methods $C(M_1, M_2)$ (e.g., *C(NSGA-II, GDE3)*) represents the percentage of the solutions in $M_2$ that are dominated by at least one solution in $M_1$.

#### *3.6.2 Statistical tests*

To test whether NSGA-II significantly outperforms GDE3, we performed two statistical hypothesis tests, namely, the Wilcoxon signed rank test and the sign test [60], [61]. The null hypothesis is defined as ‘no difference between the two methods,’ while the alternative hypothesis indicates ‘a significant difference between the two methods.’ The significance level is set as 0.05.

Also, we have differentiated between the outcomes of the NSGA-II and GDE3 algorithms by evaluating the mean, standard deviation, and coefficient of variation for the average normalized objective functions values associated with each algorithm.

# 4. Experimental settings

*A- Algorithm parameters*

For NSGA-II, the population size $m$ is set to be 100. Crossover and mutation rates are set to commonly used values of 0.8 and 0.05, respectively[14], [15].

On the other hand, for GDE3, we used a grid-search hyper-parameter tuning technique to choose the method parameters according to the number of Pareto solutions, the largest hyper volume, and the minimal fitness values, as illustrated in Table 5. Fig. 2 shows that the first experimental configuration gives the best results. Consequently, we set the scaling factor $(K)$ to be 0.5, the crossover $(C_r)$ to be 0.2, and the mutation probability $(F)$ to be 0.2.
Both NSGA-II and GDE3 algorithms were executed for 50 independent runs. Each run had a maximum of 60,000 iterations, and a set of Pareto solutions was generated.

*B- Experimental environment*

The computational environment had the following specifications*: Intel(R) Core (TM) i7-8750H CPU @ 2.20GHz-2.21 GHz, RAM (16.0 GB), Windows 10 Pro, 64-bit operating system.*

*Table 5 GDE3 hyperparameter tuning results*

| Experimental Configuration | $K$ | $C_r$ | $F$ | FV | HV | No. of Pareto-Solutions |
|---|---|---|---|---|---|---|
| 0 | 0.5 | 0.2 | 0.2 | 5,336,868 | 58,866,490 | 34 |
| 1 | 0.5 | 0.5 | 0.2 | 5,286,437 | 51,926,040 | 13 |
| 2 | 0.5 | 0.8 | 0.2 | 5,380,311 | 38,966,042 | 7 |
| 3 | 0.5 | 0.2 | 0.5 | 5,324,835 | 55,950,620 | 16 |
| 4 | 0.5 | 0.5 | 0.5 | 5,339,064 | 39,073,431 | 10 |
| 5 | 0.5 | 0.8 | 0.5 | 5,789,819 | 0 | 1 |
| 6 | 0.5 | 0.2 | 0.8 | 5,368,005 | 58,778,257 | 17 |
| 7 | 0.5 | 0.5 | 0.8 | 5,354,323 | 43,970,431 | 11 |
| 8 | 0.5 | 0.8 | 0.8 | 5,353,349 | 3,176,080 | 1 |

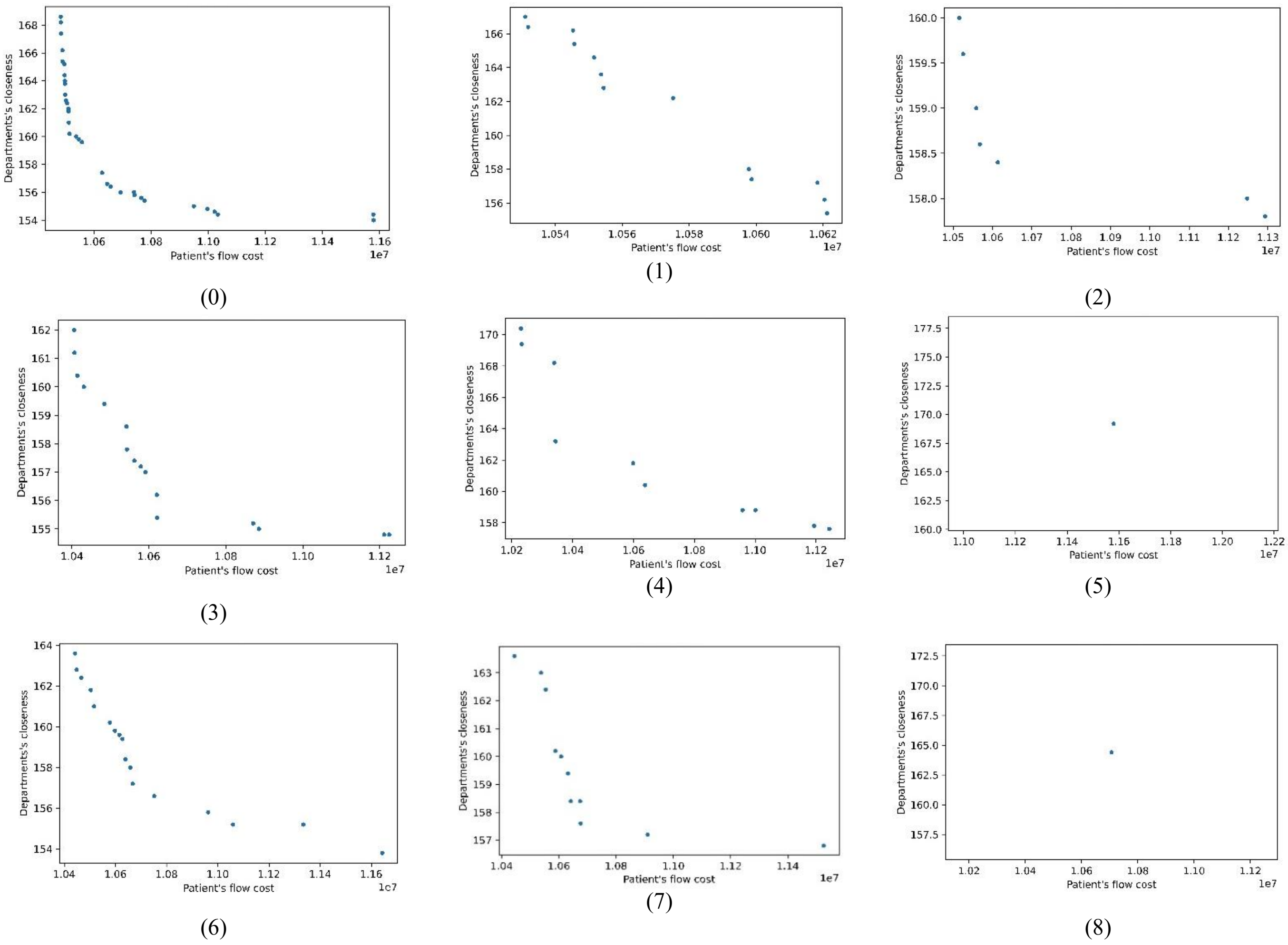

*Figure 2 Pareto fronts for nine configurations of hyperparameter settings for the GDE3 optimization method.*

In this work, to find the optimal solutions for our EDLP, we used NSGA-II and GDE3 implementations in the Python-based jMetalPy framework for multi-objective optimization with meta-heuristics [62]. This framework provides visualization modules to display the Pareto fronts with different numbers of objectives.

# 5. Results & Discussion

## 5.1. *Results of ED layout optimization*

For each of the GDE3 and NSGA-II algorithms, 50 independent runs were made, where each run produced a set of Pareto solutions. All layout solutions produced by the two methods are shown in Fig. 3. Clearly, the Pareto solutions produced by NSGA-II are better than the GDE3 ones. Each Pareto solution produced by GDE3 and NSGA-II is better than the original one in terms of the patient flow and closeness objectives, as shown in Table 6. Specifically, the layouts created by GDE3 (*resp.* NSGA-II) reduce the average flow cost by 14.19% (*resp.* 16.07%) and reduce the average closeness by 14.53% (*resp.* 15.77%), compared to the original layout. The minimum flow cost of GDE3 is smaller than the NSGA-II one, while the minimum closeness of NSGA-II is smaller than that of GDE3.

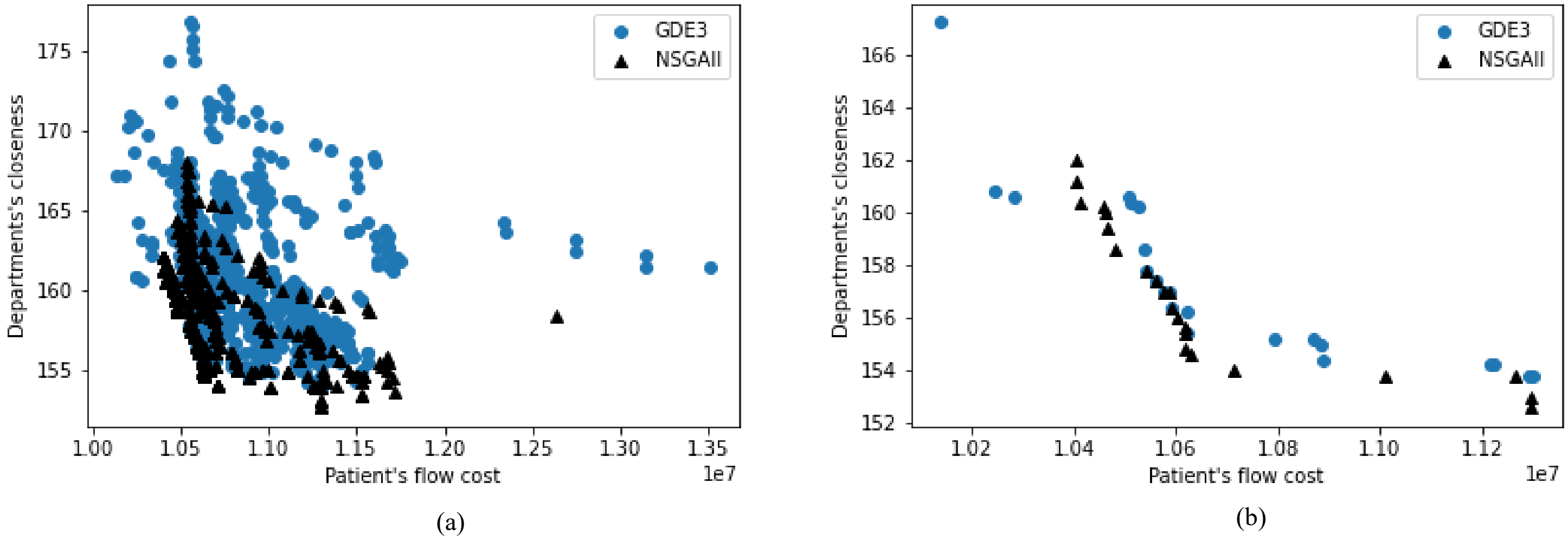

*Figure 3 Facility layout solutions: (a) Solutions with GDE3 and NSGA-II, (b) Pareto solutions with GDE3 and NSGA-II*

*Table 6 Comparison of layout solutions of NSGA-II and GDE3 methods*

| | No. of solutions | Flow cost ($F_1$) | | No. of solutions | Closeness ($F_2$) | |
|---|---|---|---|---|---|---|
| | | Average | Minimum | | Average | Minimum |
| Original Layout | 1 | 12,746,000 | 12,746,000 | 1 | 187.6 | 187.6 |
| NSGA-II | 926 | 10,697,600 | 10,402,800 | 926 | 158.019 | 152.6 |
| GDE3 | 561 | 10,937,200 | 10,138,100 | 561 | 160.343 | 153.8 |

We computed performance metrics and conducted non-parametric statistical tests as mentioned in Section 3.6 to compare between the GDE3 and NSGA-II algorithms. The average values for HV, FV, C-metric and computational time are listed in Table 7, while the results of the sign test and the Wilcoxon sign rank are illustrated in Table 8.

*Table 7 Performance measures for the NSGA-II and GDE3 methods*

| | **NSGA-II** | **GDE3** |
|---|---|---|
| **Average HV** | 56,811,250.54 | 41,500,945.32 |
| **Average FV** | 5,348,867.34 | 5,468,699.24 |
| **Average C-metric** | C (NSGA-II, GDE3) = 83.96% | C (GDE3, NSGA-II) = 6.1% |
| **Average computation time** | 3,418.73 | 3,358.3 |

*Table 8 Statistical comparison between the solutions obtained by the NSGA-II and GDE3 methods.*

| **Metrics** | **NSGA-II versus GDE3** | | | | |
|---|---|---|---|---|---|
| | **Wins** | **Equals** | **Loses** | **Sign Test (p-value)** | **Wilcoxon Sign Rank Test (p-value)** |
| **HV** | 45 | 0 | 5 | <0.0001 | <0.0001 |
| **FV** | 41 | 0 | 9 | <0.0001 | <0.0001 |
| **C-metric** | 45 | 1 | 4 | <0.0001 | <0.0001 |

We can deduce that NSGA-II has better quality of solutions than GDE3 as it has higher average HV across the 50 runs. As well, NSGA-II has the minimal average FV (and this indicates a better average distribution across both objective functions). We found that the c-metric C(NSGA-II, GDE3) > C(GDE3, NSGA-II), and this indicates that more solutions of GDE3 are dominated by those of NSGA-II. As well, the computational cost of the NSGA-II is worse than that of the GDE3.

On the other hand, Table 8 shows that both the sign test and the sign-rank test produce *p*-values well below 0.05. Thus, we can reject the null hypothesis and accept the alternate hypothesis, that there is a significant difference between NSGA-II and GDE3.

*Table 9 Statistical metrics of the normalized objective function values for the NSGA-II and GDE3 Pareto solutions.*

| | **Flow cost ($F_1$)** | **Closeness ($F_2$)** | **Mean** | **Std** | **CV** |
|---|---|---|---|---|---|
| **NSGA-II** | 0.292 | 0.454 | 0.373 | 0.114 | 0.306 |
| **GDE3** | 0.481 | 0.264 | 0.373 | 0.153 | 0.411 |

The NSGA-II and GDE3 solutions are evaluated and reported in Table 9. Clearly, the coefficient of variation for the NSGA-II solutions is smaller than that of the GDE3 ones. That is, the NSGA-II solutions show more consistency (but less diversity) than the GDE3 ones.

### 5.2. *Layout evaluation with graph measures*

As mentioned in Section 3.4, we explored local and global theoretic features of ED layout graphs (where each graph node indicates a service area while an edge reflects a (weighted) connection between two service areas. Here, a layout adjacency matrix indicates which pairs of service areas are adjacent to each other. A weight matrix associates each adjacency relation with some weight value.

*Strategies evaluation results*

As mentioned in Section 3.4, the adjacency matrix can be calculated according to different strategies as presented in Table 10 and Table 11.

*Table 10 Evaluation results of all adjacency matrix calculation strategies for different graph measures of NSGA-II solutions.*

| **NSGAII** | | | | | | |
|---|---|---|---|---|---|---|
| **Strategy** | **GE** | **NCPL** | **Transitivity** | **Mean** | **Std** | **CV** |
| **S(I), ($\gamma$ = 0)** | 0.6394 | 0.4784 | 0.4699 | 0.5292 | 0.0955 | 0.1804 |
| **S(II), ($\gamma$ = 0.25)** | 0.5912 | 0.4264 | 0.3438 | 0.4538 | 0.126 | 0.2776 |
| **S(III), ($\gamma$ = 0.5)** | 0.4606 | 0.3527 | 0.4913 | 0.4349 | 0.0728 | 0.1674 |
| **S(IV), ($\gamma$ = 0.75)** | 0.4214 | 0.4039 | 0.3953 | 0.4069 | 0.0133 | 0.0326 |
| **S(V), ($\gamma$ = 1)** | 0.4517 | 0.4641 | 0.528 | 0.4813 | 0.041 | 0.0851 |

*Table 11 Evaluation results of all adjacency matrix calculation strategies for different graph measures of GDE3 solutions.*

| GDE3 | | | | | | |
|---|---|---|---|---|---|---|
| **Strategy** | **GE** | **NCPL** | **Transitivity** | **Mean** | **Std** | **CV** |
| **S(I), (γ = 0)** | 0.5298 | 0.391 | 0.4307 | 0.4505 | 0.0715 | 0.1587 |
| **S(II), (γ = 0.25)** | 0.4804 | 0.3172 | 0.4166 | 0.4047 | 0.0823 | 0.2032 |
| **S(III), (γ = 0.5)** | 0.351 | 0.3114 | 0.4671 | 0.3765 | 0.0809 | 0.2149 |
| **S(IV), (γ = 0.75)** | 0.3536 | 0.3481 | 0.569 | 0.4236 | 0.126 | 0.2974 |
| **S(V), (γ = 1)** | 0.4241 | 0.412 | 0.4947 | 0.4436 | 0.0447 | 0.1007 |

For the NSGA-II solutions, Strategy (IV) has the lowest coefficient of variation and would thus fit the graph measures for the Pareto solutions. For the GDE3 solutions, Strategy (V) is more suitable for graph measure computations.

### *5.2.1 Graph-theoretic evaluation of Pareto solutions from NSGA-II & GDE3*

As mentioned above, we obtained Pareto solutions (layouts) using the NSGA-II and GDE3 methods. After that, we explored three types of global graph-theoretic measures to make a holistic comparison between all these Pareto solutions versus the original layout values. That is, the best solutions are selected based on the transitivity, global efficiency, and NCPL metrics. Fig. 4 shows histograms of these three measures over all Pareto solutions for the NSGA-II and GDE3 methods, respectively. Table 12 shows the average and maximum values for each of the three graph measures and each of the two optimization methods. For each of these selected solutions, the associated objective functions (i.e., the patient flow and closeness rating objectives) are inspected. Specifically, the obtained layouts are checked against the pairwise closeness ratings of the ED service areas. Consequently, we obtained the best solution for each of the NSGA-II and GDE3 methods, along with the corresponding local graph-theoretic measures.

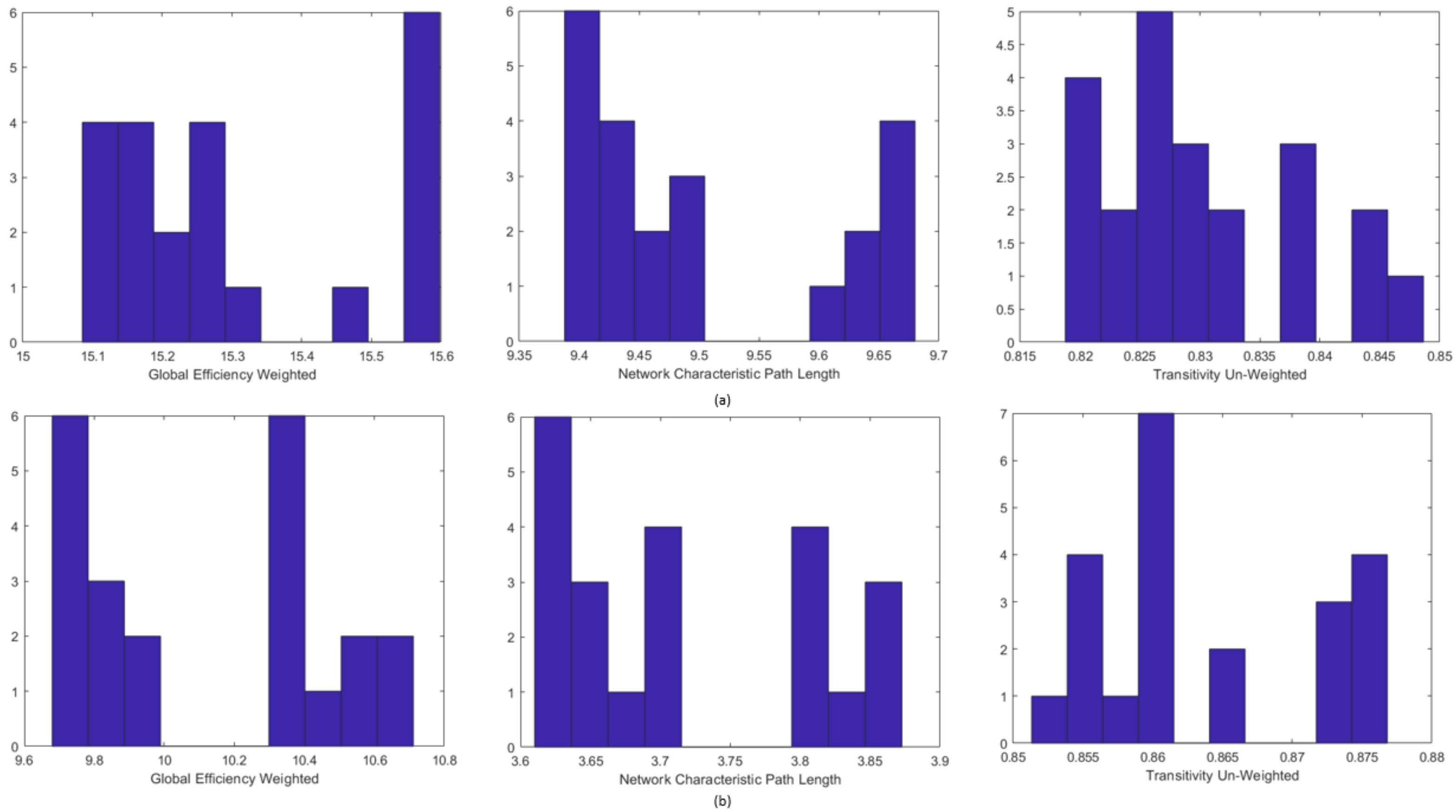

*Figure 4 Global measures for all Pareto solutions of (a) NSGA-II, and (b) GDE3.*

The objective function values corresponding to the best NSGA-II and GDE3 layouts are listed in Table 13. Clearly, although the patient flow cost of Layout A is lower than that of Layout B, the closeness for Layout B is better than that of Layout A. While there is a clear tradeoff between the two layouts in terms of the two objective functions, both layouts are better than that of the original layout. The arrangements of the service areas in Layout A and Layout B are shown in Figs. 5 and 6, respectively. Moreover, distributions of the local graph measures for Layout A and Layout B are shown in Fig. 7. First of all, the two layouts have different degree distributions (the histogram intersection distance is 0.65 as shown in Table 14). Also, the two layouts are different from the original layout in terms of the degree distribution. Layouts A and B have different distributions of the closeness centrality (CC) (with a histogram intersection distance of 0.65). The normalized betweenness centrality (BC) shows slight differences in the distribution across the two layouts (the histogram intersection distance is 0.05). While clustering coefficient shows obvious difference in the pattern and the distribution across the two layouts. The two layouts have different nodal strength distribution (histogram intersection distance is 0.6). Also, the eccentricity (ECC) seems different in pattern but may be close in the distance value for each service area. The pairwise histogram intersection distances between the layouts are illustrated in Table 14.

*Table 12 Graph metric results across whole Pareto solutions of both NSGA-II and GDE3 versus the original layout.*

| Measures \ Algorithm | Original Layout | NSGA-II | | GDE3 | |
|---|---|---|---|---|---|
| | | Average | Maximum | Average | Maximum |
| Global Efficiency (GE) | 15.5996 | 15.3012 | 15.5977 | 10.1172 | 10.7116 |
| Network Char. Path Length (NCPL) | 9.8041 | **9.5057** | 9.6800 | 3.7182 | 3.8726 |
| Transitivity | 0.8149 | 0.8306 | 0.8486 | 0.8639 | 0.8768 |

*Table 13 Objective function values for the optimized layouts versus the original layout*

| Objective Functions | Layout (A) NSGA-II | Layout (B) GDE3 | Original Layout |
|---|---|---|---|
| Patients' flow cost ($F_1$) | **10,481,348.25** | 11,289,915.471 | 12,746,000 |
| Department Closeness ($F_2$) | 158.6 | **153.8** | 187.6 |

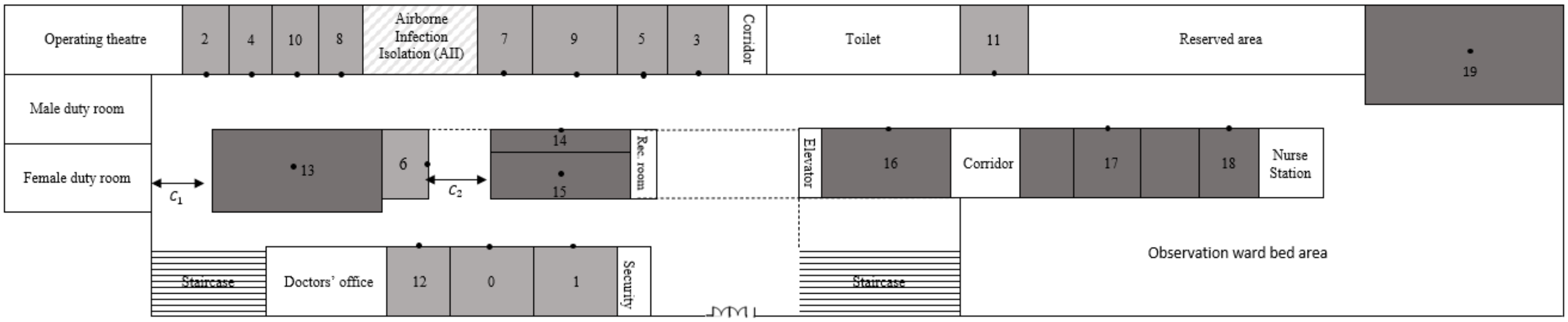

*Figure 5 Layout (A) obtained by NSGA-II, $C_1$=3.99 m, $C_2$=3.6 m*

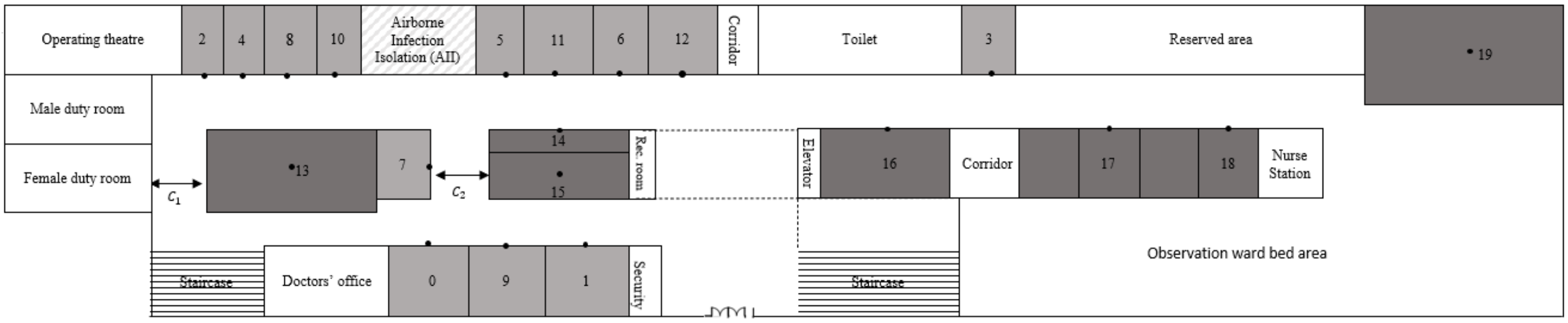

*Figure 6 Layout (B) obtained by GDE3, $C_1$=3.321 m, $C_2$=3.6 m*

*Table 14 Pairwise histogram intersection distances between layouts*

| | Original layout -Layout A | Original layout -Layout B | Layout A-Layout B |
|---|---|---|---|
| **Degree** | 0.45 | 0.8 | 0.65 |
| **Normalized BC** | 0.2 | 0.2 | 0.05 |
| **Clustering Coeff** | 0.5 | 0.35 | 0.35 |
| **Closeness centrality** | 0.25 | 0.85 | 0.65 |
| **Nodal Strength** | 0.15 | 0.75 | 0.6 |

#### *5.2.2. Local graph-theoretic measures of the original layout.*

All results of the local graph measures are presented in Fig.7. Fig. 7(a, b) for original layout, show that 14 nodes (out of 20 nodes) have a degree range between 12 to 16. Additionally, node number 16, which represents the registration service area, has the highest degree, indicating its centrality within the network. The closeness centrality results in Fig. 7(c) show that about 80 % of the nodes have closeness centrality values between 0.6 and 0.8. The betweenness centrality varies widely among nodes, from a minimum value of 0 to a maximum value of 0.1 (See Fig. 7(d)). Furthermore, node 10, representing the clinical laboratory, has the highest betweenness centrality value, indicating its involvement in a significant number of shortest paths within the network. Moreover, Fig. 7(e) shows that 85% of the nodes have a clustering coefficient between 0.7 and 1.1, 15% have a clustering coefficient between 0.4 and 0.6. The average clustering coefficient is 0.7974, indicating that the original layout represents a high aggregation network. Also, Fig. 7(f) shows that the network has maximum and minimum nodal strengths of 500 and 100, respectively, with the nodal strengths for 85% of the nodes being between 100 and 300. In Fig. 7(g), the eccentricity measure shows a maximum distance between nodes 15 (triage (1)) and 20 (radiology).

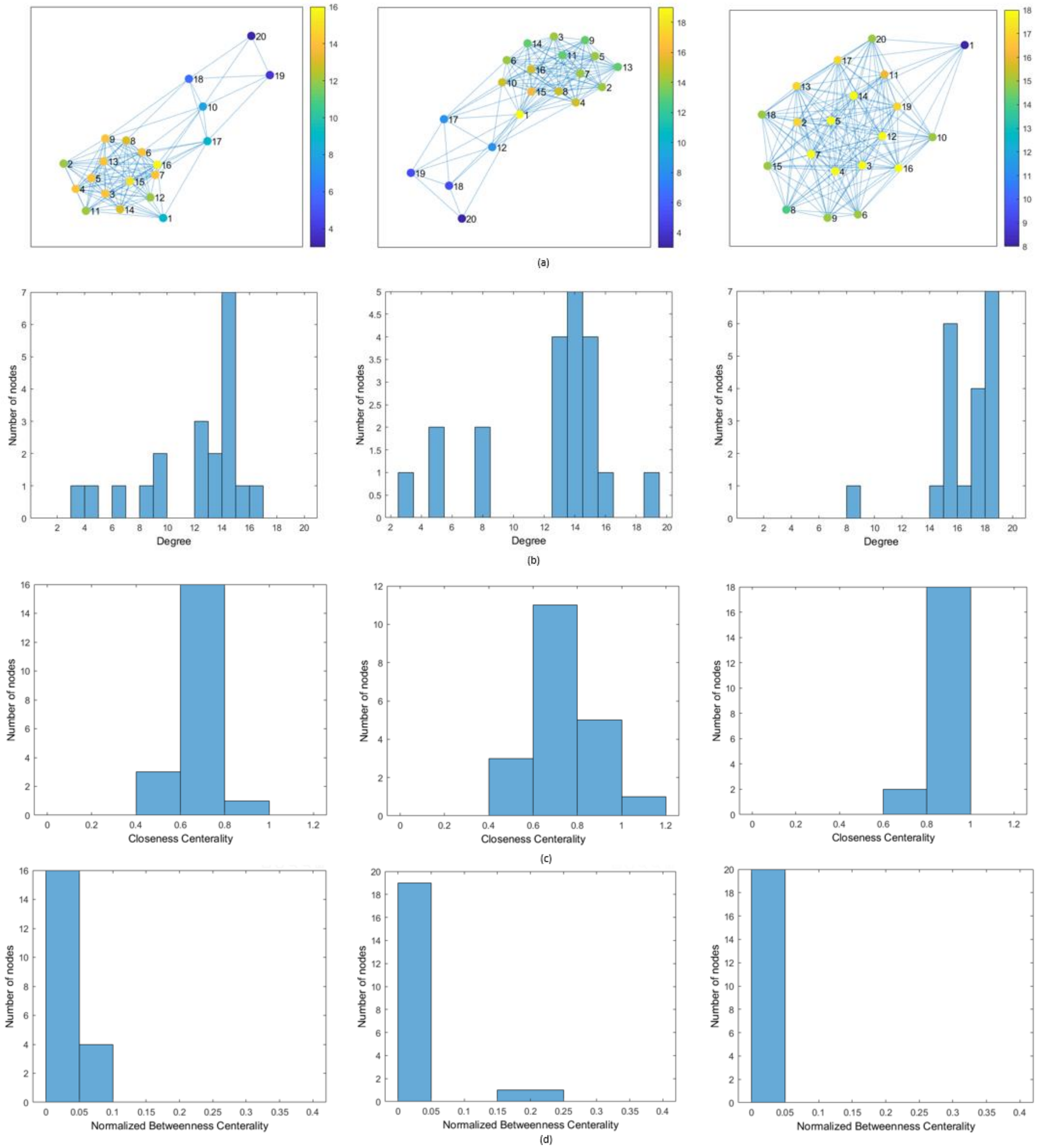

*Figure 7 Local graph measures for the original layout, Layout A, and Layout B (shown in the 1st, 2nd, and 3rd column, respectively). (a) degree distribution, (b) degree, (c) Closeness centrality and (d) Normalized BC.*

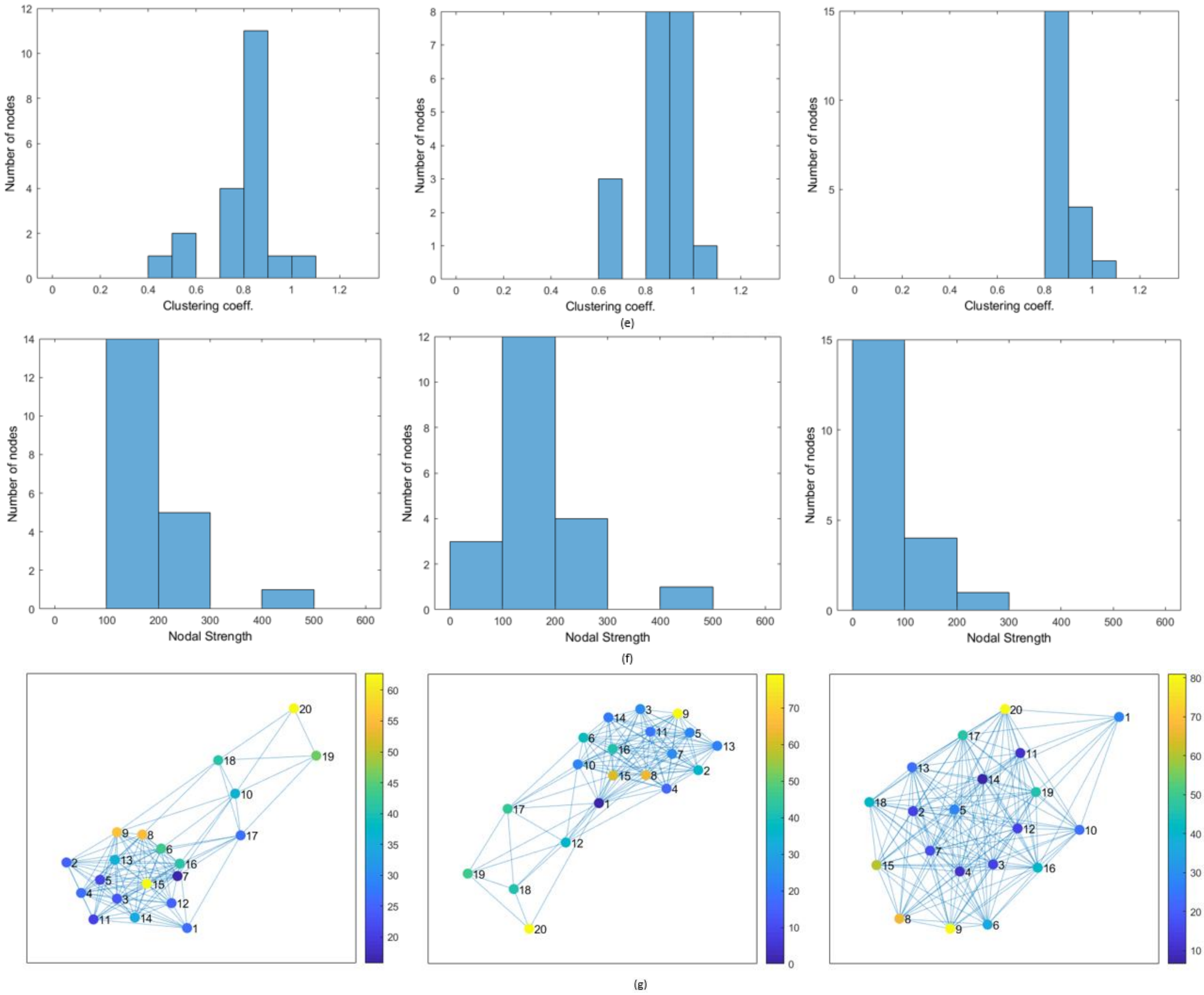

*Continue Figure 7: Local graph measures for the original layout, Layout A, and Layout B (shown in the 1st, 2nd, and 3rd column, respectively). (e) Clustering Coeff, (f) Nodal strength, and (g) ECC*

### 5.3. ***Results of Pareto Solution Ranking using the ELECTRE-III Method***

We experimented with different methods for ranking the obtained Pareto solutions: ranking by the objective function values, ranking by global graph-theoretic measures, and ranking by combinations of both function value and graph-theoretic measures. The results for each of these three ranking approaches are shown next.

### *5.3.2. Ranking by objective function values*

First of all, the optimal Pareto solutions obtained from both the NSGA-II and GDE3 methods were ranked by the objective function value following the ELECTRE-III technique. The ranking results for NSGA-II and GDE3 are shown in Table 15. We found that the solutions $S_1$ and $S_{19}$ have the same rank across all NSGA-II solutions, while $S_{17}$ obtained by GDE3 is the best. The layouts for the $S_1$, $S_{19}$, and $S_{17}$ solutions are visualized in Fig. 8. Each of these layouts can be compactly represented as follows. For the NSGA-II-based $S_1$, the orders of the service areas in the first, second, and third rows can be listed as {[2,4,10,8,5,9,7,11,3], [6], [0,12,1]}, while the corridor widths are $C_1 = 3.431\ m$ and $C_2 = 3.6\ m$. The corresponding representation for $S_{19}$ is {[2,4,10,8,7,11,12,6,3], [5], [0,9,1]}, $C_1 = 3.273\ m$, and $C_2 = 3.6\ m$. For the GDE3-based $S_{17}$, the representation is {[2,4,8,10,5,11,6,12,3], [7], [9,0,1]}, $C_1 = 3.273\ m$ and $C_2 = 3.6\ m$.

*Table 15 Ranking of the NSGA-II and GDE3 solutions by the objective function values.*

| | 1 | 2 | 3 | 4 | 5 | 6 | 7 | 8 | 9 | 10 | 11 | 12 | 13 | 14 |
|---|---|---|---|---|---|---|---|---|---|---|---|---|---|---|
| **NSGA-II Solutions** | $S_1$, $S_{19}$ | $S_3$, $S_7$, $S_{22}$ | $S_{17}$ | $S_{16}$ | $S_9$ | $S_{18}$ | $S_{12}$, $S_5$, $S_{11}$, $S_{14}$, $S_{15}$ | $S_{10}$ | $S_{13}$, $S_{21}$ | $S_{20}$ | $S_2$, $S_4$, $S_6$ | $S_8$ | - | - |
| **GDE3-Solutions** | $S_{17}$ | $S_{11}$ | $S_{10}$, $S_{18}$, $S_{22}$ | $S_{14}$ | $S_9$, $S_{19}$ | $S_{20}$ | $S_{12}$, $S_3$, $S_4$ | $S_8$ | $S_{16}$ | $S_1$ | $S_{13}$ | $S_2$, $S_5$, $S_6$, $S_{21}$ | $S_7$ | $S_{15}$ |

### *5.3.2 Ranking by global graph measures*

The Pareto solutions were also ranked by the global graph measures in order to assess the suitability of these measures in obtaining plausible and meaningful solutions. Tables 16 shows the ranking results for the NSGA-II and GDE3 solutions. Clearly, six of the NSGA-II solutions had the same top ranking (where S6 is Layout A, and the rest having small differences in their representations). For the GDE3 solutions, $S_{15}$ (i.e., Layout B) ranked at the top.

*Table 16 Ranking of NSGA-II and GDE3 solutions by global graph measures*

| | 1 | 2 | 3 | 4 | 5 | 6 | 7 |
|---|---|---|---|---|---|---|---|
| **NSGA-II Solutions** | $S_5$, $S_6$, $S_7$, $S_{11}$, $S_{14}$, $S_{15}$, | $S_{18}$ | $S_2$ | $S_9$, $S_{16}$, $S_{17}$, $S_{19}$, $S_{22}$ | $S_1$, $S_3$, $S_4$, $S_8$, $S_{12}$, $S_{13}$, $S_{20}$, $S_{21}$ | $S_{10}$ | - |
| **GDE3-Solutions** | $S_{15}$ | $S_7$, $S_{14}$, $S_{17}$ | $S_{18}$ | $S_3$, $S_5$, $S_9$, $S_{10}$, $S_{19}$, $S_{20}$ | $S_1$ | $S_2$, $S_4$, $S_6$, $S_8$, $S_{12}$, $S_{13}$, $S_{16}$, $S_{21}$, $S_{22}$ | $S_{11}$ |

### *5.3.2 Ranking by combined objective function values and graph-theoretic measures*

Finally, we ranked the solutions according to the combination of global graph measures and objective functions. The NSGA-II and GDE3 ranking results are shown in Tables 17. Clearly, the top-ranking NSGA-II solution is $S_{19}$ (see Figure 8(b)), while the top-ranking GDE3 one is $S_{10}$ (where the representation for $S_{10}$ is [2,4,10,8,5,9,6,11,3], [7], [0,12,1], $C_1$=

3.4307 m and $C_2$=3.6 m).

*Table 17 Ranking of NSGA-II solutions according to both global GM & objective fn.*

| | **1** | **2** | **3** | **4** | **5** | **6** | **7** | **8** | **9** | **10** | **11** | **12** | **13** | **14** |
|---|---|---|---|---|---|---|---|---|---|---|---|---|---|---|
| **NSGA-II Solutions** | $S_{19}$ | $S_1$, $S_3$, $S_7$, $S_{22}$ | $S_{16}$, $S_{17}$ | $S_9$ | $S_{18}$ | $S_{12}$, $S_5$, $S_{11}$, $S_{14}$, $S_{15}$ | $S_{20}$ | $S_{13}$, $S_{21}$ | $S_2$, $S_4$, $S_6$, $S_{10}$ | $S_8$ | - | - | - | - |
| **GDE3 Solutions** | $S_{10}$ | $S_{18}$ | $S_{17}$ | $S_{14}$ | $S_{11}$, $S_{22}$ | $S_{20}$ | $S_9$ | $S_{19}$ | $S_{12}$, $S_3$, $S_4$ | $S_7$, $S_{13}$ | $S_8$ | $S_1$, $S_{16}$ | $S_6$, $S_{15}$ | $S_2$, $S_5$, $S_{21}$ |

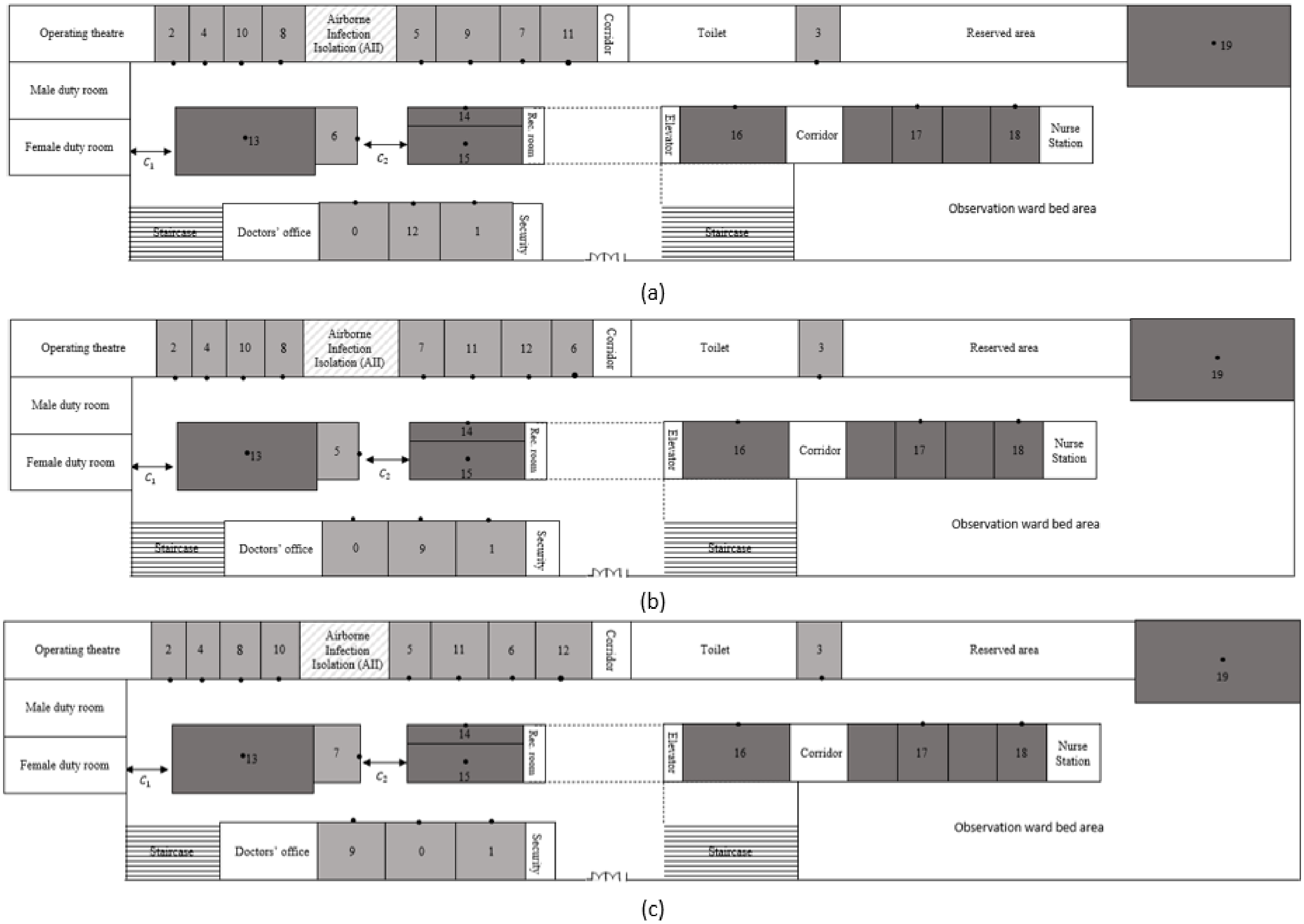

*Figure 8 Optimal layouts after ranking the Pareto solutions by the objective function values using the ELECTRE-III method: (a) $S_1$ from NSGA-II, (b) $S_{19}$ from NSGA-II, and (c) $S_{17}$ from GDE3.*

As shown in Table 18, we compare the optimal solutions obtained by the ranking approaches and choose the best overall solution which should have low objective functions values but high values of graph-theoretic measures.

*Table 18 Comparison between the original layout and the optimal solutions obtained by the three ranking approaches.*

| | Solutions ranked by objective functions | | | Solutions ranked by graph-theoretic measures | | Solutions ranked by combined objective functions and graph-theoretic measures | | Original Layout | Original Layout |
|---|---|---|---|---|---|---|---|---|---|
| | $S_1$ NSGA-II | $S_{19}$ NSGA-II | $S_{17}$ GDE3 | $S_6$ NSGA-II (Layout A) | $S_{15}$ GDE3 (Layout B) | $S_{19}$ NSGA-II | $S_{10}$ GDE3 | ($\gamma = 1$) | ($\gamma = 0.75$) |
| Patients' flow cost ($F_1$) | 10,587,896 | 11,263,771 | 11,297,511 | 10,411,205.93 | 11,289,915.47 | 11,263,771 | 10,883,149.60 | 12,746,000 | 12,746,000 |
| Closeness ($F_2$) | 157 | 153.8 | 153.8 | 160.4 | 153.8 | 153.8 | 155 | 187.6 | 187.6 |
| Global Efficiency | 15.096 | 15.2337 | 10.596 | 15.5532 | 10.7116 | 15.2337 | 10.323 | 11.2057 | 15.5996 |
| NCPL | 9.3899 | 9.5002 | 3.8278 | 9.6581 | 3.8177 | 9.5002 | 3.7048 | 3.9931 | 9.8041 |
| Transitivity | 0.8188 | 0.8269 | 0.8768 | 0.839 | 0.8768 | 0.8269 | 0.872 | 0.8591 | 0.8149 |

We can see that the NSGA-II solution $S_6$ (Layout A) has the lowest patient flow cost and the highest graph-theoretic measures (but still its closeness function value is worse than the others). The GDE3 solution $S_{10}$ is worse than $S_{15}$ and $S_{17}$ (except for the patient flow cost). The solution $S_{15}$ (Layout B) is better than $S_{17}$ (except for the NCPL measure). Collectively, the best solutions obtained through ranking by the ELECTRE III method, namely, $S_6$ (Layout A) and $S_{15}$ (Layout B), are indeed the solutions with the best graph-theoretic measures (as mentioned in Section 5.2.2). In comparison with the original layout, Layout A (the optimal NSGA-II solution) reduces $F_1$ by 18.32 % and improves $F_2$ by 14.5 %. Also, Layout B (the optimal GDE3 solution) reduces $F_1$ by 11.42 % and improves $F_2$ by 18.02 %.

We can conclude that the NSGA-II and GDE3 techniques optimize both patients' flow cost ($F_1$) and closeness ($F_2$ ) and this indicates the need for some adjustment of the layout of the service areas (getting some to be close to each other while keeping others apart). We observe that the optimal Layout A agrees with the closeness ratings listed in Table 4.

# 6. Conclusions and Future Work

In this paper, we formulated a real EDLP as a meta-heuristic multi-objective optimization problem. The NSGA-II and GDE3 global optimization techniques were adopted to solve this problem. Experiments showed that NSGA-II can find better Pareto solutions than GDE3. In particular, sets of Pareto optimal ED layouts were sought to achieve minimal patient flow and maximal closeness between the ED service areas. Then, we considered each ED layout as a graph from which we computed graph-theoretic measures. Then, a multi-criteria decision-making technique was employed to rank the candidate layouts according to different criteria including the patient flow and closeness objective values, the introduced graph measures, or combinations of both types. Indeed, the graph-theoretic measures appeared to be useful in enhancing the interpretability of the Pareto solutions, and hence these measures can help medical planning experts in selecting the best layouts. Overall, we recommend using such graph-theoretic measures as novel quantitative measures for evaluating healthcare facility layouts.

Future research directions may include exploring the hospital layout problem in the context of alternative design structures. This future research can build upon the current work by introducing additional constraints that reflect the problem's complexity in actual clinical settings. For example, designing a multi-floor hospital would necessitate addressing the unique challenges posed by vertical space and vertical transportation [15]. As well, we can consider the formulated problem to be a multi-objective dynamic ED layout problem by which patient flow can be analyzed towards enhanced layouts [16]. Also, other global optimization techniques can be investigated such as Swarm-based optimization technique [63]. Furthermore, we would explore integrating the agent-based and discrete-event simulation approach to enhance service quality especially the length of stay for the patient [64].

# Declarations

**Availability of data and material:** All layout, flow, and closeness rating data are available in Zuo *et al.* [14], doi: 10.1109/TASE.2018.2873098.

**Competing Interests:** The authors declare no known financial or personal conflicts that could have influenced the research presented in this paper.

**Funding:** The authors declare that no funds, grants, or other support were received during the preparation of this manuscript.

**Authors' contributions:** All authors contributed to the study conception and design. Data analysis and the first draft were completed by O.S. Revision of the first draft and supervision of the work were done by M.R. and M.A. All authors read and approved the final manuscript.

**Acknowledgements:** Not applicable.

# Appendix A

**<u>Algorithm I: NSGA-II Algorithm</u>**

**Inputs:**

- Number of evaluations (stopping criteria)
- NSGA-II parameters (population size (*m* =100), crossover and mutation rates are set to be 0.8 and 0.05 respectively)
- Objective functions to be optimized

Generation of initial population (*P*)
Evaluating the objective functions
Ranking (non-dominated sorting and crowding distance)
Generate offspring population of size (*m*)
**for** *i* =1: stopping criteria **do**
    **for** each parent and offspring ϵ P **do**
        Ranking
        Generating sets of non-dominated solutions (Selection)

9. Crossover and mutation
10. Loop based on existing solution to get next generation
11. Evaluating the objective functions
12. **end for**
13. Select points on the lower front with high distance
14. Generate next generation
15. **end for**

**Output:** Pareto front (set of non-dominated solutions from final population)

**Algorithm II: GDE3 Algorithm**

**Inputs:**

- Number of evaluations (stopping criteria)
- GDE3 parameters (population size ($m$ =100), crossover rate ($C_r$), mutation rate ($F$) and scaling factor ($K$) are set to be 0.2, 0.2 and 0.5 respectively)
- Objective functions to be optimized

1. Generation of initial population (*P*)
2. Evaluating the objective functions
3. Repeat for generation g = 1 to stopping criteria
4. **for** *i* =1: stopping criteria **do**
5. **for** individual ϵ *P* **do**
6. Mutation:
   i. (Randomly select three distinct individuals ($x_1$, $x_2$, $x_3$) from P (excluding i)
   ii. Create mutant vector $v_i$:
   $v_{ij} = x_{1j} + F * (x_{2j} - x_{3j})$for each dimension j
   Apply random uniform value between 0 and 1 to each dimension j with probability $F$. If the value is less than $F$, replace $v_{ij}$ with the corresponding value from individual $i$. (This ensures some elements remain unchanged)

7. Crossover:
   i. Initialize trial vector $u_i$
   ii. For each dimension $j$:
   Random uniform value between 0 and 1 (*rand_j*)
   If (*rand_j* < or *j* = *j_rand*), set $u_{ij} = v_{ij}$ (*j_rand* is a randomly chosen index)
   Otherwise, set $u_{ij}$ = target vector component from individual $i$ (based on a selection strategy, e.g., randomly chosen)
8. Selection:
   i. Evaluate objectives functions for trial vector $u_i$
   ii. If $u_i$ dominates individual $i$ or both are non-dominated and $u_i$ has better crowding distance; replace individual $i$ with $u_i$ in the population.

**Output:** Pareto front (set of non-dominated solutions)

**<u>Algorithm III: ELECTRE-III Algorithm</u>**

1. Define alternatives & criteria.
   a. Alternatives are Pareto solutions.
   b. Criteria are conventional objective functions, global graph measures, or the combination of conventional and graph-based criteria.
2. Assign weights to criteria, ($W_i = 1$)
3. Set indifference threshold ($q_i <= 15\%$) , preference threshold ($p_i = 30\%$), and veto threshold ($v_i = 30\%$), where $q_i \leq p_i \leq v_i$ is verified.

4. Calculate pairwise concordance $\left(c_i\ (a_m, a_n)\right)$ & discordance $\left(d_i\ (a_m, a_n)\right)$

$$c_i\ (a_m, a_n) = \begin{cases} 0, & p_i < g_i(a_m) - g_i(a_n) \\ \dfrac{g(a_m) + p_i - g(a_n)}{p_i - q_i}, & q_i < g_i(a_n) - g_i(a_m) \leq p_i \\ 1, & g_i(a_n) - g_i(a_m) \leq q_i \end{cases}$$

$$d_i\ (a_m, a_n) = \begin{cases} 0, & g_i(a_n) - g_i(a_m) < p_i \\ \dfrac{g(a_m) + p_i - g(a_n)}{p_i - q_i}, & p_i \leq g_i(a_n) - g_i(a_m) \leq v_i \\ 1, & v_i < g_i(a_n) - g_i(a_m) \end{cases}$$

5. Check for veto & build dominance matrix.
6. Rank alternatives based on dominance.
7. Check if all alternatives are ranked.